\documentclass{article} 
\usepackage{iclr2025_conference,times}

\usepackage{amsmath,amsfonts,bm}

\def\eqref#1{equation~\ref{#1}}

\def\1{\bm{1}}

\DeclareMathAlphabet{\mathsfit}{\encodingdefault}{\sfdefault}{m}{sl}
\SetMathAlphabet{\mathsfit}{bold}{\encodingdefault}{\sfdefault}{bx}{n}

\usepackage{hyperref}
\usepackage{url}
\usepackage{float}
\usepackage{graphicx} 
\usepackage{tabularx}
\usepackage{booktabs}
\usepackage{tabularx}
\usepackage{pgfplots}
\pgfplotsset{compat=1.17}
\usepackage{caption}
\usepackage{graphicx}
\usepackage{subcaption}

\title{Auto-ABSA: Cross-Domain Aspect Detection and Sentiment Analysis Using Auxiliary Sentences}
\iclrfinalcopy 

\author{Teng Wang\\
Department of Mathematics, The University of Hong Kong, Hong Kong SAR, China\\
\texttt{wt0318@connect.hku.hk} \\
\AND
Bolun Sun \\
Johns Hopkins University, Baltimore, MD, USA  \\
Address \\
\texttt{bsun26@jh.edu}\\
\AND
Yijie Tong  \\
Southern University of Science and Technology, Shenzhen, Guangzhou, China \\
\texttt{11811512@mail.sustech.edu.cn} \\
}

\begin{document}

\maketitle

\begin{abstract}
Since the introduction of the transformer architecture, numerous pre-trained language models have been developed, significantly advancing the sentiment analysis (SA) task. In this paper, we propose an innovative approach that utilizes auxiliary sentences to describe the aspects present in a given sentence, thereby enhancing sentiment prediction. The proposed method consists of two main components. First, aspect detection is performed using a multi-aspect detection model to identify all the aspects contained in a sentence. These predicted aspects are then combined with the original sentence as input to the sentiment analysis model. Second, we conduct out-of-domain aspect-based sentiment analysis (ABSA) by training a sentiment classification model on one type of dataset and validating it on a different dataset to evaluate its generalizability. To benchmark our approach, we established two baselines: one without aspect information and another using all available aspects as input for sentiment classification. By comparing the performance of these baselines with our method, we demonstrate that our approach effectively improves sentiment prediction. The results highlight the potential benefits of incorporating auxiliary aspect information, contributing to the development of more robust SA models.
\end{abstract}

\section{Introduction}
With the rapid advancement of information technology, users now have access to an abundance of information and services. Various fields have leveraged information technology to employ natural language processing (NLP) techniques for extracting valuable insights from massive unstructured text data. Sentiment analysis (SA), a subtask of NLP, focuses on identifying and extracting sentiments expressed in text. It can be used to analyze user reviews of products, assess public opinion on news, and even predict election outcomes.

Predicting the sentiment polarity of a sentence with a single sentiment is relatively straightforward. However, sentences often convey multiple sentiments related to different aspects. For instance, a sentence may contain multiple targets, each with distinct sentiments for different aspects. Consider the example: "I love using the iPhone because of its quality, but it's too expensive." In this review, the sentiment regarding price is negative, while the sentiment regarding quality is positive. Aspect-based sentiment analysis (ABSA), a subtask of SA, aims to determine the sentiment polarity for each aspect of a given target. Unlike simple text classification, ABSA requires models to accurately detect the sentiment associated with different aspects, making it a more complex and challenging task.

Several previous studies have attempted to address this challenge. For example, Tang et al.\cite{tang2015effective} proposed a model using two LSTM neural networks: one network processes preceding contexts along with the target, while the other processes following contexts and the target. The outputs of the two LSTMs are then combined, and a softmax function is applied to predict the sentiment polarity. However, LSTMs have limitations, such as lower performance compared to attention-based mechanisms and a lack of parallelizability. Sun et al.\cite{sun2019utilizing} leveraged BERT along with auxiliary sentences to predict sentiment polarity, while Wan et al.\cite{wan2020target} proposed a method involving full permutations of aspects and polarities, represented as "auxiliary sentences," to predict sentiment. However, this approach has two main drawbacks: it requires prior knowledge of all possible aspects within a dataset, limiting its cross-domain applicability, and suffers from data imbalance due to a large number of auxiliary sentences.

Gao et al.\cite{gao2019target} proposed a method using auxiliary sentences as prior knowledge to determine sentiment polarities (positive, negative, neutral), but it does not explicitly consider aspects, which can be problematic when one target corresponds to multiple aspects. He et al.\cite{he2020deberta} introduced the DeBERTa model, employing disentangled attention mechanisms, and demonstrated state-of-the-art performance in ABSA tasks.

When NLP models are deployed in new domains, it often becomes necessary to re-analyze the data to adapt to domain-specific characteristics. This involves significant effort in feature extraction and requires large amounts of high-quality labeled data, which can be expensive and time-consuming to obtain. Cross-domain aspect-based sentiment analysis aims to address these challenges by utilizing models trained on annotated data from a source domain to analyze unlabeled data in a target domain. For instance, training on restaurant reviews and testing on real estate reviews represents a challenging cross-domain setting.

Words are crucial in sentiment classification tasks, and their sentiment orientation may vary across domains. For instance, the term "unpredictable" might be perceived as positive in movie reviews but negative when referring to electronic components. Cross-domain ABSA seeks to overcome these discrepancies and utilize existing models trained in one domain to efficiently perform tasks in other domains. Automatic detection of aspects in ABSA is one of the most effective approaches for achieving this objective.

In this paper, we propose a cross-domain sentiment analysis model based on deep learning methods. First, we evaluate the performance of the sentiment classification model when given either all aspects present in the dataset or none as part of the model input. Next, we provide the correct aspects from the SemEval16 dataset to a sentiment classification model trained on the Sentihood dataset. Our experiments show that incorporating the correct aspects significantly improves the model's performance compared to two baseline settings. To obtain accurate aspect information, we adapt SpanEmo\cite{alhuzali2021spanemo}, a model originally designed for multi-emotion detection, to predict aspects in sentences. By combining SpanEmo with a sentiment predictor, we introduce a comprehensive "Big Model" framework, named Auto-ABSA, which performs cross-domain ABSA effectively.

Our contributions are summarized as follows:
\begin{enumerate}
    \item We modify SpanEmo to enable multi-aspect detection for ABSA.
    \item We train the model on one type of dataset and validate it on a different type, demonstrating the effectiveness of using auxiliary sentences for cross-domain ABSA.
    \item We enhance model interpretability by examining the influence of aspect information on sentiment predictions. Through comparison with two baseline settings, we demonstrate that our approach significantly improves sentiment classification accuracy.
\end{enumerate}

\section{Related Work}
In early studies, sentiment classification was primarily built using models such as LSTM \cite{hochreiter1997long}, GRU \cite{chung2014empirical}, etc. Tang et al.\cite{tang2015effective} combined two LSTM neural networks: one network processed preceding contexts along with the target, while the other processed following contexts and the target. Wang et al.\cite{wang2016attention} proposed an attention-based LSTM to predict polarity at the aspect level. Chen et al.\cite{chen2017recurrent} introduced a model utilizing a Bi-RNN, location-weighted memory, and attention mechanisms. However, these models did not achieve satisfactory performance.

After the transformer architecture was introduced by Vaswani et al.\cite{vaswani2017attention}, attention mechanisms \cite{bahdanau2014neural} gained widespread popularity, resulting in the development of numerous pre-trained models, advanced attention mechanisms, and novel training strategies. Notable models include BERT \cite{devlin2018bert}, RoBERTa \cite{liu2019roberta}, and DeBERTa \cite{he2020deberta}. Gao et al.\cite{gao2019target} and Sun et al.\cite{sun2019utilizing} proposed methods using BERT and auxiliary sentences as prior knowledge to enhance sentiment prediction. Wan et al.\cite{wan2020target} employed BERT to construct a multi-task model that predicts both the polarity of aspects and the target's location at the word level. However, this method relies on all permutations of aspect polarities (positive, negative, neutral) and aspects present in the dataset, resulting in data imbalance and a model prone to outputting 0, making it challenging to train effectively on sentiment prediction tasks.

\begin{figure}[H]
\centering
\includegraphics[width=0.5 \textwidth]{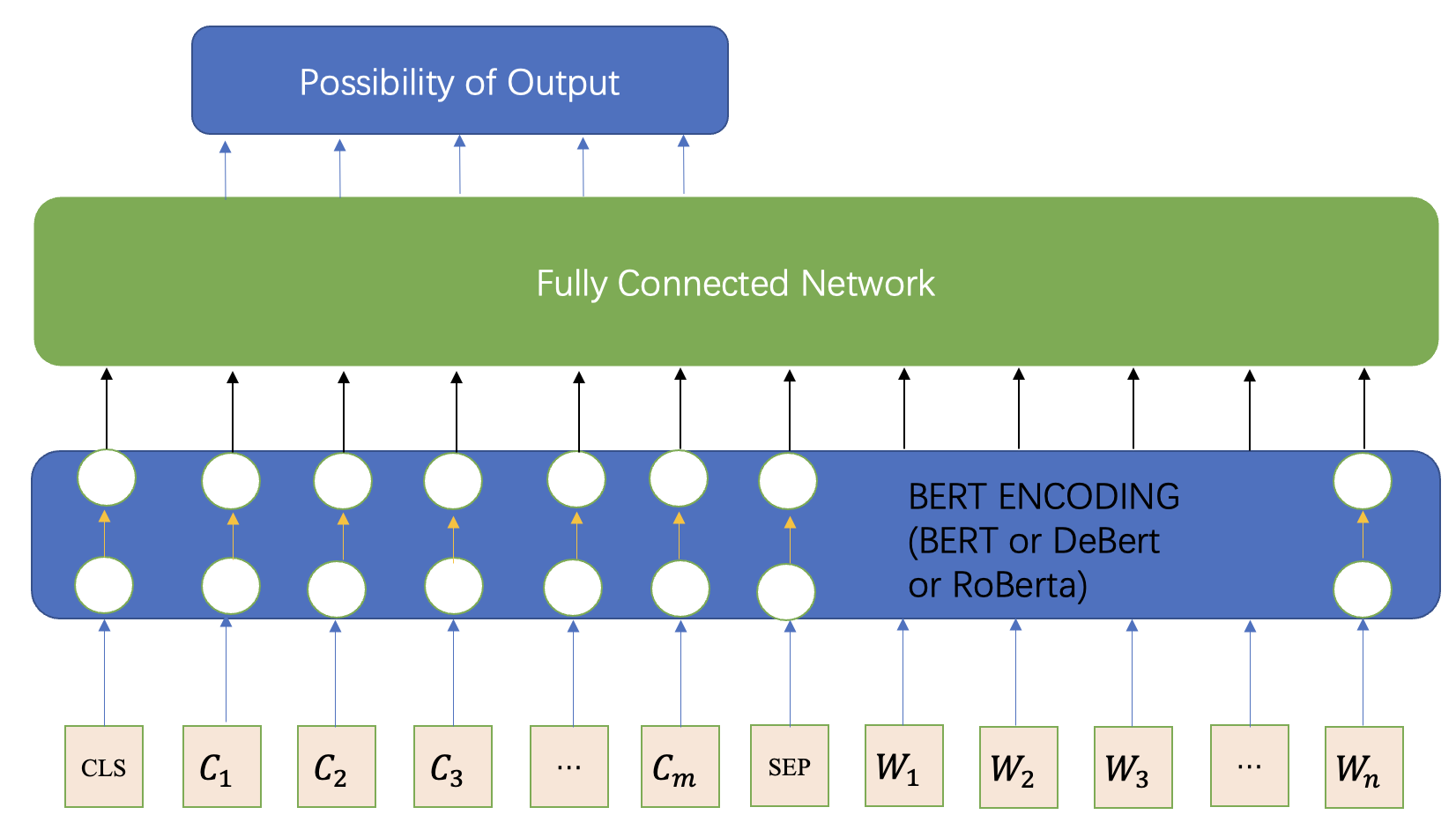}
\caption{\label{fig:SpanEmo} The architecture of the SpanEmo model.}
\end{figure}

Our aspect detection model builds upon SpanEmo \cite{alhuzali2021spanemo}, which was initially designed to predict multiple emotions within a sentence. We adapt SpanEmo to predict aspects instead of emotions, evaluating its capability to accurately identify all aspects present in a sentence. Figure \ref{fig:SpanEmo} illustrates the SpanEmo architecture. The input to the model is a combination of all aspects and the given sentence, and the output represents the likelihood of each aspect being present.

\section{Methodology}

\subsection{SpanEmo}

Let $\{S, Y\}$ denote a dataset comprising $N$ examples, where $S = \{s_1, s_2, \dots, s_N\}$ is the set of sentences, and $Y = \{y_1, y_2, \dots, y_N\}$ is the set of corresponding aspect labels. Each sentence $s_i$ is associated with an aspect label vector $y_i \in \{0,1\}^C$, where $C$ is the number of predefined aspect classes, and each element indicates the presence ($1$) or absence ($0$) of an aspect in $s_i$.

In our proposed model, SpanEmo, the input consists of two components: the set of fixed aspect classes and the corresponding sentence. We encode the input using a BERT encoder to obtain the hidden representations $H_i \in \mathbb{R}^{L \times D}$, where $L$ is the maximum input length, and $D$ is the hidden state dimensionality:
\begin{equation}
H_i = \text{BERT}([\text{CLS}] + [C] + [\text{SEP}] + s_i)
\end{equation}
Here, $[C]$ represents the tokenized aspect classes, and $s_i$ is the tokenized sentence.

To predict all aspects present in a sentence, we formulate this as a multi-label binary classification task. We apply a Fully Connected Network (FCN) followed by a sigmoid activation function to predict the probability $\hat{y}_i$ of each aspect being present:
\begin{equation}
\hat{y}_i = \text{sigmoid}(\text{FCN}(H_i))
\end{equation}

We adopt the loss function from the original SpanEmo paper, which combines the Binary Cross-Entropy (BCE) loss and the Label-Correlation Aware (LCA) loss. The LCA loss penalizes the model for predicting pairs of labels that should not co-occur, capturing label dependencies:
\begin{equation}
L_{\text{LCA}}(y_i, \hat{y}_i) = \frac{1}{|y_i^0||y_i^1|} \sum_{(p,q) \in y_i^0 \times y_i^1} \exp(\hat{y}_{i,p} - \hat{y}_{i,q})
\end{equation}
where $y_i^0$ and $y_i^1$ are the sets of negative and positive labels for the $i$-th example, respectively, and $\hat{y}_{i,p}$ is the $p$-th element of $\hat{y}_i$.

The total loss function is a weighted sum of the BCE loss and the LCA loss:
\begin{equation}
L = (1 - \alpha) L_{\text{BCE}} + \alpha \sum_{i=1}^N L_{\text{LCA}}(y_i, \hat{y}_i)
\end{equation}
where $\alpha$ is a hyperparameter balancing the two loss components.

Our approach extends the original SpanEmo model by integrating it into an out-of-domain sentiment analysis framework, enabling the identification and classification of aspects in sentences from unseen domains.

\subsection{Sentiment Predictor}

\begin{figure}[H]
\centering
\includegraphics[width=0.5\textwidth]{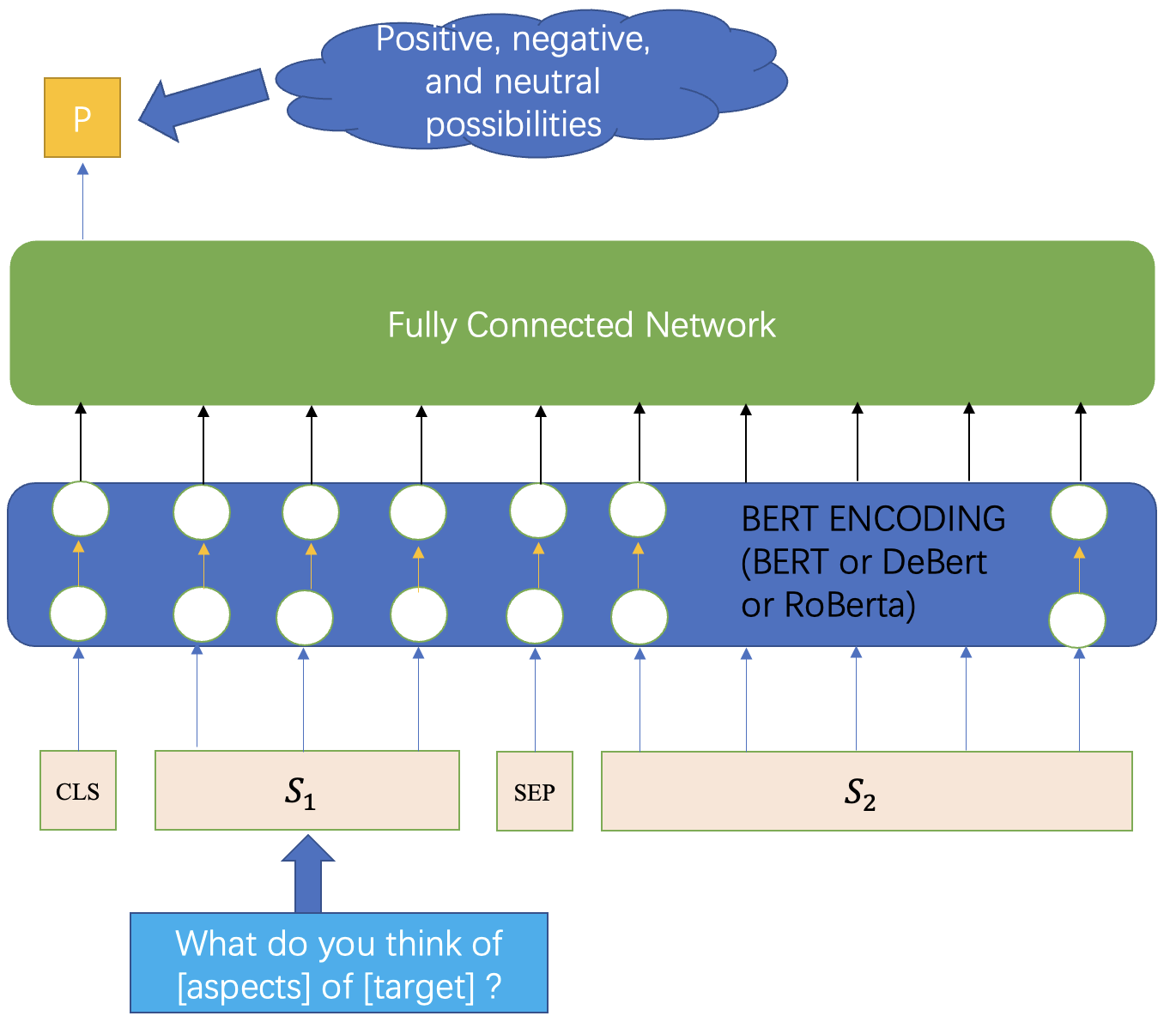}
\caption{\label{fig:sentimentPredictor} The architecture of the sentiment predictor model.}
\end{figure}

Following the construction of SpanEmo, we develop a sentiment predictor designed to generalize across domains, allowing us to predict sentiment even when there is a domain shift between the training and testing datasets (e.g., training on urban life data and testing on restaurant reviews). The architecture of the sentiment predictor is illustrated in Figure \ref{fig:sentimentPredictor}.

Our sentiment predictor assesses the transferability of sentiment models trained on one domain to another, which is crucial for practical applications where domain-specific labeled data may be scarce. This capability enables us to perform out-of-domain aspect-based sentiment analysis (ABSA) using three different methods.

We propose the following methods:
\begin{enumerate}
    \item \textbf{SpanEmo + Sentiment Predictor}: We first use SpanEmo to identify the aspects present in a sentence. The predicted aspects, along with the target and the sentence, are then input to the sentiment predictor, which outputs the probabilities of positive, negative, and neutral sentiment.
    \item \textbf{Sentiment Predictor with All Aspects}: We directly input all possible aspects, the target, and the sentence into the sentiment predictor, bypassing the aspect identification step.
    \item \textbf{Sentiment Predictor without Aspects}: We input the target and the sentence into the sentiment predictor without any aspect information.
\end{enumerate}

To illustrate these methods, consider the example sentence: "The food is delicious, but it's too expensive," with aspects including \textit{price}, \textit{quality}, and \textit{atmosphere}. Table \ref{tab:input_examples} presents the input formats for each method.

\begin{table}[H]
\centering
\renewcommand{\arraystretch}{1.5}
\caption{Examples of input formats for each method}
\label{tab:input_examples}
\begin{tabularx}{\textwidth}{|X|X|}
\hline
\textbf{Method} & \textbf{Input} \\
\hline
SpanEmo + Sentiment Predictor & [CLS] What do you think of quality and price of the food? [SEP] The food is delicious, but it's too expensive [SEP] \\
\hline  
Sentiment Predictor with All Aspects & [CLS] What do you think of quality, price, and atmosphere of the food? [SEP] The food is delicious, but it's too expensive [SEP] \\
\hline
Sentiment Predictor without Aspects & [CLS] What do you think of NULL of the food? [SEP] The food is delicious, but it's too expensive [SEP] \\
\hline
\end{tabularx}
\end{table}

We define our dataset as $\{(A_i, T_i, S_i)\}_{i=1}^N$, where $A_i$ denotes the aspects present in the $i$-th sentence, $T_i$ denotes the target, and $S_i$ denotes the sentence itself. The input representation for each example is formulated as:
\begin{equation}
\text{input}_i = [\text{CLS}] + s_{1,i} + [\text{SEP}] + s_{2,i}
\end{equation}
where $s_{1,i}$ is the prompt "What do you think of [aspects] of [target]?" and $s_{2,i}$ is the original sentence $S_i$.

After encoding the input with BERT, we obtain the hidden representation $H_i$. To perform sentiment classification, we apply a Fully Connected Network followed by a Softmax activation function to obtain the probabilities $\hat{y}_i \in \mathbb{R}^{3}$ of the positive, negative, and neutral sentiment classes:
\begin{equation}
H_i = \text{BERT}(\text{input}_i)
\end{equation}
\begin{equation}
\hat{y}_i = \text{Softmax}(\text{FCN}(H_i))
\end{equation}

The loss function for the sentiment predictor is the cross-entropy loss:
\begin{equation}
L_i = -\sum_{c=1}^3 y_{i,c} \log(\hat{y}_{i,c})
\end{equation}
where $y_{i,c}$ is a binary indicator (1 if the true class label is $c$, 0 otherwise), and $\hat{y}_{i,c}$ is the predicted probability that the $i$-th observation belongs to class $c$.

\subsection{Out-of-Domain ABSA}

Our objective is to perform out-of-domain aspect-based sentiment analysis (ABSA), aiming to make predictions on datasets for which no specific training data is available. By integrating SpanEmo and the sentiment predictor, we propose a novel method for out-of-domain ABSA capable of performing aspect extraction and sentiment classification without requiring domain-specific training data.

\begin{figure}[H]
\centering
\includegraphics[width=0.5\textwidth]{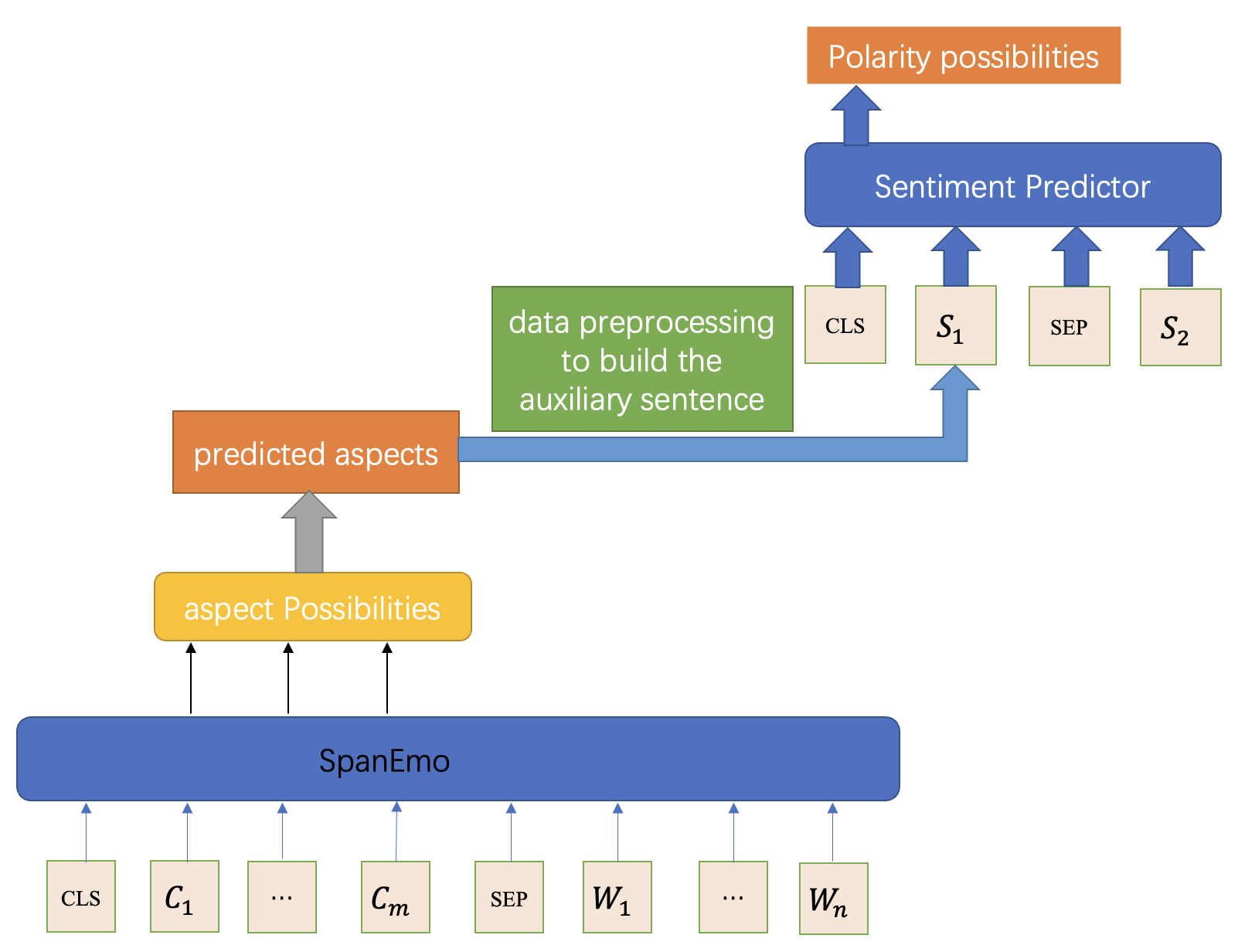}
\caption{\label{fig:BigModel} The architecture of the Big Model combining SpanEmo and the sentiment predictor for out-of-domain ABSA.}
\end{figure}

The \textbf{Big Model}, depicted in Figure \ref{fig:BigModel}, combines SpanEmo and the sentiment predictor. We first train SpanEmo on a suitable dataset to predict aspects in new sentences. The predicted aspects, along with the target and the sentence, are then input to the sentiment predictor, which outputs the sentiment probabilities for the target.

Alternatively, if all possible aspects of the sentence are known, we can use the sentiment predictor directly with all aspects (method 2). If aspect information is unavailable, we can use the sentiment predictor without aspects by providing NULL as the aspect input (method 3).

These methods enable us to perform out-of-domain ABSA, demonstrating the flexibility and generalizability of our approach across different domains.

\section{Experiments}

In this section, we present a comprehensive evaluation of our proposed model, including the datasets utilized, training configurations, and experimental results. All implementations were carried out using PyTorch \cite{paszke2019pytorch}, and experiments were conducted on an NVIDIA V100 GPU with 16 GB memory. We fine-tuned and compared three pre-trained language models: BERT-base, DeBERTa-base, and RoBERTa-base.

\subsection{Datasets}

We employed four publicly available datasets for training, evaluation, and testing:

\begin{itemize}
    \item \textbf{SentiHood}: An urban neighborhood reviews dataset annotated with aspect categories and sentiment polarities.
    \item \textbf{SemEval-2014 Task 4}: A benchmark dataset for aspect-based sentiment analysis in restaurant reviews.
    \item \textbf{SemEval-2015 Task 12}: An extension focusing on more fine-grained aspect categories.
    \item \textbf{SemEval-2016 Task 5}: A dataset targeting aspect-based sentiment analysis in restaurant and laptop domains.
\end{itemize}

These datasets provide a diverse set of domains and aspect categories, enabling us to evaluate the generalizability of our model across different contexts.

\subsection{Training Details}

For training \textbf{SpanEmo}, we partitioned each dataset into 80\% for training, 10\% for validation, and 10\% for testing. Through extensive hyperparameter tuning, we identified optimal settings listed in Table~\ref{SpanEmoHyper}.

As an illustration, consider the SemEval-2016 dataset, which contains six aspect categories: \textit{food}, \textit{restaurant}, \textit{atmosphere}, \textit{drinks}, \textit{location}, and \textit{service}. For example, given the sentence:

\begin{quote}
``After all that, they complained to me about the small tip.''
\end{quote}

The relevant aspect is \textit{service}. The input to the model is constructed as:

\begin{quote}
$[\text{CLS}]$ \textit{food, restaurant, atmosphere, drinks, location, service} $[\text{SEP}]$ \textit{After all that, they complained to me about the small tip.}
\end{quote}

The model predicts probabilities for each aspect, assigning a higher probability to \textit{service} (exceeding a predefined threshold), while assigning lower probabilities to the other aspects.

For training the \textbf{Sentiment Predictor}, we similarly partitioned the data into 80\% for training, 10\% for validation, and 10\% for testing. Since SpanEmo predicts aspects for a single target, the Sentiment Predictor is designed to predict the sentiment polarity for one target at a time. Therefore, sentences containing multiple targets were excluded from the training set. The optimal hyperparameters determined through experimentation are listed in Table~\ref{SentimentPredictorHyper}.

We constructed the input to the Sentiment Predictor by generating an auxiliary sentence incorporating the relevant aspects and target, formatted as:

\begin{quote}
``What do you think of [aspects] of [target]?''
\end{quote}

This auxiliary sentence is concatenated with the original sentence to form the input. The Sentiment Predictor then outputs the probabilities for each sentiment polarity class.

\begin{table}[H]
\centering
\begin{tabularx}{\textwidth}{lX}  
\toprule
\textbf{Hyperparameter}      & \textbf{Value}          \\
\midrule
Maximum Sequence Length      & 128                     \\
Dropout Rate                 & 0.2                     \\
Training Batch Size          & 32                      \\
Validation Batch Size        & 32                      \\
BERT Learning Rate           & $2 \times 10^{-5}$      \\
FCN Learning Rate            & $1 \times 10^{-3}$      \\
Alpha (Loss Weight)          & 0.2                     \\
Optimizer                    & Adam                    \\
\bottomrule
\end{tabularx}
\caption{Hyperparameters for Training SpanEmo}
\label{SpanEmoHyper}
\end{table}

\begin{table}[H]
\centering
\begin{tabularx}{\textwidth}{lX}
\toprule
\textbf{Hyperparameter}      & \textbf{Value}          \\
\midrule
Weight Decay                 & Linear                  \\
Warm-up Steps                & 10\% of Total Steps     \\
Maximum Sequence Length      & 128                     \\
Dropout Rate                 & 0.1                     \\
Training Batch Size          & 32                      \\
Validation Batch Size        & 32                      \\
BERT Learning Rate           & $2 \times 10^{-5}$      \\
FCN Learning Rate            & $1 \times 10^{-3}$      \\
Optimizer                    & Adam                    \\
\bottomrule
\end{tabularx}
\caption{Hyperparameters for Training Sentiment Predictor}
\label{SentimentPredictorHyper}
\end{table}

\subsection{Experiment I: Aspect Detection}

To evaluate the performance of SpanEmo on aspect detection, we trained and tested the model separately on each dataset. The results are summarized in Tables~\ref{table:spanemo_semeval2014}, \ref{table:spanemo_semeval2015}, and \ref{table:spanemo_semeval2016}. Performance metrics include macro- and micro-averaged F1 scores, Jaccard similarity (JS), precision, and recall.

\begin{table}[H]
\centering
\begin{tabular}{lccccc}
\toprule
\textbf{Model} & \textbf{F1-Macro} & \textbf{F1-Micro} & \textbf{JS} & \textbf{Precision} & \textbf{Recall} \\
\midrule
BERT-base      & 0.7551            & 0.6999            & 0.7126      & 0.7551             & 0.7171          \\
DeBERTa-base   & 0.7371            & 0.7129            & 0.6726      & 0.7911             & 0.6899          \\
RoBERTa-base   & \textbf{0.7752}   & \textbf{0.7707}   & \textbf{0.7519} & \textbf{0.7752} & \textbf{0.7752} \\
\bottomrule
\end{tabular}
\caption{Performance of SpanEmo on SemEval-2014 Dataset}
\label{table:spanemo_semeval2014}
\end{table}

\begin{table}[H]
\centering
\begin{tabular}{lccccc}
\toprule
\textbf{Model} & \textbf{F1-Macro} & \textbf{F1-Micro} & \textbf{JS} & \textbf{Precision} & \textbf{Recall} \\
\midrule
BERT-base      & 0.7000            & 0.4660            & 0.6750      & 0.7000             & 0.7000          \\
DeBERTa-base   & \textbf{0.7347}   & \textbf{0.4191}   & \textbf{0.6687} & \textbf{0.8060} & \textbf{0.6750} \\
RoBERTa-base   & 0.5862            & 0.2786            & 0.5312      & 0.6538             & 0.5312          \\
\bottomrule
\end{tabular}
\caption{Performance of SpanEmo on SemEval-2015 Dataset}
\label{table:spanemo_semeval2015}
\end{table}

\begin{table}[H]
\centering
\begin{tabular}{lccccc}
\toprule
\textbf{Model} & \textbf{F1-Macro} & \textbf{F1-Micro} & \textbf{JS} & \textbf{Precision} & \textbf{Recall} \\
\midrule
BERT-base      & 0.8213            & 0.5307            & 0.7748      & 0.8438             & 0.8000          \\
DeBERTa-base   & \textbf{0.8401}   & \textbf{0.6424}   & \textbf{0.8168} & 0.8433          & \textbf{0.8370} \\
RoBERTa-base   & 0.8333            & 0.5364            & 0.8333      & \textbf{0.8527}    & 0.8148          \\
\bottomrule
\end{tabular}
\caption{Performance of SpanEmo on SemEval-2016 Dataset}
\label{table:spanemo_semeval2016}
\end{table}

The results demonstrate that SpanEmo effectively identifies aspects across different datasets, with RoBERTa-base achieving the highest F1 scores on SemEval-2014 and DeBERTa-base performing best on SemEval-2016.

\subsection{Experiment II: Sentiment Prediction}

We evaluated the Sentiment Predictor's ability to generalize across domains by training it on the SentiHood dataset and testing it on the SemEval-2016 dataset. We conducted experiments under three settings:

\begin{enumerate}
    \item \textbf{With Correct Aspects}: The model is provided with the correct aspects.
    \item \textbf{With All Aspects}: The model is provided with all possible aspects.
    \item \textbf{Without Aspects}: The model is not provided with any aspect information.
\end{enumerate}

Tables~\ref{table:sentiment_correct_aspects}, \ref{table:sentiment_all_aspects}, and \ref{table:sentiment_no_aspects} present the performance under each setting.

\begin{table}[H]
\centering
\begin{tabular}{lcc}
\toprule
\textbf{Model} & \textbf{F1-Micro} & \textbf{F1-Macro} \\
\midrule
BERT-base      & \textbf{0.7734}   & \textbf{0.6330}   \\
DeBERTa-base   & 0.7750            & 0.6076            \\
RoBERTa-base   & 0.7508            & 0.6026            \\
\bottomrule
\end{tabular}
\caption{Sentiment Prediction with Correct Aspects}
\label{table:sentiment_correct_aspects}
\end{table}

\begin{table}[H]
\centering
\begin{tabular}{lcc}
\toprule
\textbf{Model} & \textbf{F1-Micro} & \textbf{F1-Macro} \\
\midrule
BERT-base      & 0.6586            & 0.2647            \\
DeBERTa-base   & \textbf{0.7453}   & \textbf{0.2847}   \\
RoBERTa-base   & 0.6586            & 0.2647            \\
\bottomrule
\end{tabular}
\caption{Sentiment Prediction with All Aspects}
\label{table:sentiment_all_aspects}
\end{table}

\begin{table}[H]
\centering
\begin{tabular}{lcc}
\toprule
\textbf{Model} & \textbf{F1-Micro} & \textbf{F1-Macro} \\
\midrule
BERT-base      & \textbf{0.6648}   & \textbf{0.2662}   \\
DeBERTa-base   & 0.6531            & 0.2634            \\
RoBERTa-base   & 0.6630            & 0.2659            \\
\bottomrule
\end{tabular}
\caption{Sentiment Prediction without Aspects}
\label{table:sentiment_no_aspects}
\end{table}

The results indicate that providing the correct aspects significantly improves sentiment prediction performance. The Sentiment Predictor achieves higher F1 scores when supplied with accurate aspect information, highlighting the importance of aspect knowledge in cross-domain sentiment analysis.

\subsection{Experiment III: Out-of-Domain Aspect-Based Sentiment Analysis}

To assess the effectiveness of our approach in out-of-domain settings, we combined SpanEmo (trained on SemEval-2016) with the Sentiment Predictor (trained on SentiHood). We refer to this integrated model as the \textbf{Big Model}. SpanEmo predicts aspects in the target domain, and the Sentiment Predictor utilizes these predicted aspects for sentiment classification.

Tables~\ref{table:bigmodel_semeval2016}, \ref{table:bigmodel_semeval2014}, and \ref{table:bigmodel_semeval2015} present the performance of the Big Model on SemEval-2016, SemEval-2014, and SemEval-2015 datasets, respectively.

\begin{table}[H]
\centering
\begin{tabular}{lcc}
\toprule
\textbf{Model} & \textbf{F1-Micro} & \textbf{F1-Macro} \\
\midrule
BERT-base      & 0.7391            & 0.6174            \\
DeBERTa-base   & 0.7203            & 0.5909            \\
RoBERTa-base   & \textbf{0.7648}   & \textbf{0.6261}   \\
\bottomrule
\end{tabular}
\caption{Performance of Big Model on SemEval-2016 Dataset}
\label{table:bigmodel_semeval2016}
\end{table}

\begin{table}[H]
\centering
\begin{tabular}{lcc}
\toprule
\textbf{Model} & \textbf{F1-Micro} & \textbf{F1-Macro} \\
\midrule
BERT-base      & 0.6622            & 0.5965            \\
DeBERTa-base   & \textbf{0.7200}   & \textbf{0.6082}   \\
RoBERTa-base   & 0.4400            & 0.4670            \\
\bottomrule
\end{tabular}
\caption{Performance of Big Model on SemEval-2014 Dataset}
\label{table:bigmodel_semeval2014}
\end{table}

\begin{table}[H]
\centering
\begin{tabular}{lcc}
\toprule
\textbf{Model} & \textbf{F1-Micro} & \textbf{F1-Macro} \\
\midrule
BERT-base      & \textbf{0.7000}   & \textbf{0.5350}   \\
DeBERTa-base   & 0.5000            & 0.5000            \\
RoBERTa-base   & 0.5813            & 0.4299            \\
\bottomrule
\end{tabular}
\caption{Performance of Big Model on SemEval-2015 Dataset}
\label{table:bigmodel_semeval2015}
\end{table}

Comparing these results with the baselines in Tables~\ref{table:sentiment_all_aspects} and \ref{table:sentiment_no_aspects}, we observe that the Big Model achieves superior performance. Providing predicted aspects as prior knowledge enhances the interpretability and effectiveness of the Sentiment Predictor in out-of-domain settings. This demonstrates that accurate aspect prediction is instrumental in transferring sentiment analysis models across domains.

Our approach offers a practical solution for out-of-domain ABSA without the need for domain-specific training data. When faced with a new dataset lacking annotated resources, our method can effectively perform sentiment classification by leveraging the generalization capabilities of SpanEmo and the Sentiment Predictor.

An interesting observation is that the Big Model based on RoBERTa-base sometimes outperforms the model provided with the correct aspects (as seen in Tables~\ref{table:bigmodel_semeval2016} and \ref{table:sentiment_correct_aspects}). This counterintuitive result may be attributed to the robustness of the combined model and the influence of random factors during training. Further investigation is warranted to understand this phenomenon fully.

\subsection{Qualitative Analysis}

To illustrate the practical advantages of our method, we present examples where SpanEmo accurately detects aspects, leading to correct sentiment predictions without requiring user-provided aspects. For instance, given the sentence:

\begin{quote}
``The ambiance was lovely, but the service was slow.''
\end{quote}

SpanEmo correctly identifies \textit{ambiance} and \textit{service} as aspects. The Sentiment Predictor then assigns a positive sentiment to \textit{ambiance} and a negative sentiment to \textit{service}, demonstrating the model's ability to perform nuanced sentiment analysis autonomously.

These qualitative results highlight the strength of our method in reducing the dependency on manual aspect annotation, thereby improving scalability and applicability in real-world scenarios.

In summary, our experiments validate the effectiveness of integrating SpanEmo with a sentiment predictor for out-of-domain aspect-based sentiment analysis. The proposed method achieves competitive performance across multiple datasets and pre-trained models, emphasizing its potential for practical applications where domain-specific resources are limited.

\section{Discussion}

In this study, we investigated the effectiveness of integrating SpanEmo with a sentiment predictor to perform out-of-domain aspect-based sentiment analysis (ABSA). Our experimental results across multiple datasets and pre-trained language models demonstrate the potential of our approach in generalizing sentiment analysis tasks to unseen domains.

\subsection{Analysis of Aspect Detection}

The performance of SpanEmo on aspect detection tasks, as presented in Tables~\ref{table:spanemo_semeval2014}, \ref{table:spanemo_semeval2015}, and \ref{table:spanemo_semeval2016}, indicates that the model effectively identifies aspects across different domains. Notably, RoBERTa-base achieved the highest F1 scores on the SemEval-2014 dataset, while DeBERTa-base performed best on SemEval-2015 and SemEval-2016 datasets. This suggests that the choice of pre-trained language model can significantly impact aspect detection performance.

The superior performance of DeBERTa-base on certain datasets may be attributed to its enhanced mask decoder and disentangled attention mechanisms, which better capture complex linguistic patterns relevant to aspect detection. However, the variation in performance across datasets also highlights the influence of domain-specific characteristics on model effectiveness.

\subsection{Impact of Aspect Information on Sentiment Prediction}

Our experiments on sentiment prediction under different settings (Tables~\ref{table:sentiment_correct_aspects}, \ref{table:sentiment_all_aspects}, and \ref{table:sentiment_no_aspects}) underscore the importance of accurate aspect information. Providing the correct aspects to the sentiment predictor yielded the highest F1 scores, emphasizing that aspect knowledge plays a crucial role in accurately determining sentiment polarity.

When all possible aspects were provided, the performance degraded compared to using the correct aspects, but still outperformed the setting without any aspect information. This indicates that while additional aspect information can be beneficial, irrelevant or extraneous aspects may introduce noise, potentially confusing the model.

Interestingly, even without any aspect information, the sentiment predictor achieved reasonable performance, suggesting that pre-trained language models possess some inherent ability to infer sentiment based on contextual cues alone. However, the addition of aspect information clearly enhances the model's capability to focus on relevant portions of the text, improving sentiment classification accuracy.

\subsection{Effectiveness of the Big Model in Out-of-Domain Settings}

The Big Model, which combines SpanEmo and the sentiment predictor, demonstrated strong performance in out-of-domain ABSA tasks (Tables~\ref{table:bigmodel_semeval2016}, \ref{table:bigmodel_semeval2014}, and \ref{table:bigmodel_semeval2015}). Notably, the Big Model often outperformed models provided with correct aspects, suggesting that the integrated approach may capture domain-generalizable patterns that enhance sentiment prediction.

This counterintuitive result may stem from the Big Model's ability to mitigate domain-specific biases present in the training data. By relying on predicted aspects rather than ground-truth aspects from a different domain, the model may avoid overfitting to domain-specific aspect distributions, leading to better generalization.

\subsection{Limitations and Future Work}

Despite the promising results, our approach has certain limitations. The reliance on accurate aspect prediction means that errors in SpanEmo's output can propagate to the sentiment predictor, potentially degrading performance. Additionally, while our method reduces the need for domain-specific training data, the pre-trained language models themselves may still carry domain biases learned during pre-training.

Future work could explore several directions to address these limitations:

\begin{itemize}
    \item \textbf{Domain Adaptation Techniques}: Implementing domain adaptation strategies, such as adversarial training or fine-tuning with a small amount of domain-specific data, could enhance the model's robustness to domain shifts.
    \item \textbf{Multi-Task Learning}: Simultaneously training the aspect detection and sentiment prediction tasks in a multi-task learning framework may improve overall performance by allowing shared representations.
    \item \textbf{Incorporating External Knowledge}: Leveraging external knowledge bases or ontologies could enrich the model's understanding of domain-specific aspects and sentiment expressions.
    \item \textbf{Error Analysis}: Conducting a detailed error analysis to identify common failure cases may provide insights for refining the model architecture and training procedures.
\end{itemize}

\section{Conclusion}

In this paper, we presented a novel approach for out-of-domain aspect-based sentiment analysis by integrating SpanEmo, an aspect detection model, with a sentiment predictor. Our method effectively leverages pre-trained language models to generalize sentiment analysis tasks across domains without requiring domain-specific training data.

Through extensive experiments on multiple benchmark datasets, we demonstrated that our integrated model achieves competitive performance, often surpassing models provided with correct aspects. The results highlight the importance of accurate aspect detection and the potential of our approach in practical applications where annotated resources are limited.

Our findings contribute to the ongoing efforts in developing robust, domain-independent sentiment analysis systems. By reducing the dependency on domain-specific data and manual aspect annotation, our method enhances the scalability and applicability of ABSA in real-world scenarios.

Future research directions include exploring domain adaptation techniques, incorporating multi-task learning frameworks, and integrating external knowledge sources to further improve model performance and generalizability. We believe that our work lays a foundation for more resilient sentiment analysis systems capable of adapting to the ever-evolving landscape of user-generated content.

\bibliography{iclr2025_conference}

\begin{thebibliography}{15}
\providecommand{\natexlab}[1]{#1}
\providecommand{\url}[1]{\texttt{#1}}
\expandafter\ifx\csname urlstyle\endcsname\relax
  \providecommand{\doi}[1]{doi: #1}\else
  \providecommand{\doi}{doi: \begingroup \urlstyle{rm}\Url}\fi

\bibitem[Alhuzali \& Ananiadou(2021)Alhuzali and Ananiadou]{alhuzali2021spanemo}
Hassan Alhuzali and Sophia Ananiadou.
\newblock Spanemo: Casting multi-label emotion classification as span-prediction.
\newblock \emph{arXiv preprint arXiv:2101.10038}, 2021.

\bibitem[Bahdanau et~al.(2014)Bahdanau, Cho, and Bengio]{bahdanau2014neural}
Dzmitry Bahdanau, Kyunghyun Cho, and Yoshua Bengio.
\newblock Neural machine translation by jointly learning to align and translate.
\newblock \emph{arXiv preprint arXiv:1409.0473}, 2014.

\bibitem[Chen et~al.(2017)Chen, Sun, Bing, and Yang]{chen2017recurrent}
Peng Chen, Zhongqian Sun, Lidong Bing, and Wei Yang.
\newblock Recurrent attention network on memory for aspect sentiment analysis.
\newblock In \emph{Proceedings of the 2017 conference on empirical methods in natural language processing}, pp.\  452--461, 2017.

\bibitem[Chung et~al.(2014)Chung, Gulcehre, Cho, and Bengio]{chung2014empirical}
Junyoung Chung, Caglar Gulcehre, KyungHyun Cho, and Yoshua Bengio.
\newblock Empirical evaluation of gated recurrent neural networks on sequence modeling.
\newblock \emph{arXiv preprint arXiv:1412.3555}, 2014.

\bibitem[Devlin et~al.(2018)Devlin, Chang, Lee, and Toutanova]{devlin2018bert}
Jacob Devlin, Ming-Wei Chang, Kenton Lee, and Kristina Toutanova.
\newblock Bert: Pre-training of deep bidirectional transformers for language understanding.
\newblock \emph{arXiv preprint arXiv:1810.04805}, 2018.

\bibitem[Gao et~al.(2019)Gao, Feng, Song, and Wu]{gao2019target}
Zhengjie Gao, Ao~Feng, Xinyu Song, and Xi~Wu.
\newblock Target-dependent sentiment classification with bert.
\newblock \emph{IEEE Access}, 7:\penalty0 154290--154299, 2019.

\bibitem[He et~al.(2020)He, Liu, Gao, and Chen]{he2020deberta}
Pengcheng He, Xiaodong Liu, Jianfeng Gao, and Weizhu Chen.
\newblock Deberta: Decoding-enhanced bert with disentangled attention.
\newblock \emph{arXiv preprint arXiv:2006.03654}, 2020.

\bibitem[Hochreiter \& Schmidhuber(1997)Hochreiter and Schmidhuber]{hochreiter1997long}
Sepp Hochreiter and J{\"u}rgen Schmidhuber.
\newblock Long short-term memory.
\newblock \emph{Neural computation}, 9\penalty0 (8):\penalty0 1735--1780, 1997.

\bibitem[Liu et~al.(2019)Liu, Ott, Goyal, Du, Joshi, Chen, Levy, Lewis, Zettlemoyer, and Stoyanov]{liu2019roberta}
Yinhan Liu, Myle Ott, Naman Goyal, Jingfei Du, Mandar Joshi, Danqi Chen, Omer Levy, Mike Lewis, Luke Zettlemoyer, and Veselin Stoyanov.
\newblock Roberta: A robustly optimized bert pretraining approach.
\newblock \emph{arXiv preprint arXiv:1907.11692}, 2019.

\bibitem[Paszke et~al.(2019)Paszke, Gross, Massa, Lerer, Bradbury, Chanan, Killeen, Lin, Gimelshein, Antiga, et~al.]{paszke2019pytorch}
Adam Paszke, Sam Gross, Francisco Massa, Adam Lerer, James Bradbury, Gregory Chanan, Trevor Killeen, Zeming Lin, Natalia Gimelshein, Luca Antiga, et~al.
\newblock Pytorch: An imperative style, high-performance deep learning library.
\newblock \emph{Advances in neural information processing systems}, 32:\penalty0 8026--8037, 2019.

\bibitem[Sun et~al.(2019)Sun, Huang, and Qiu]{sun2019utilizing}
Chi Sun, Luyao Huang, and Xipeng Qiu.
\newblock Utilizing bert for aspect-based sentiment analysis via constructing auxiliary sentence.
\newblock \emph{arXiv preprint arXiv:1903.09588}, 2019.

\bibitem[Tang et~al.(2015)Tang, Qin, Feng, and Liu]{tang2015effective}
Duyu Tang, Bing Qin, Xiaocheng Feng, and Ting Liu.
\newblock Effective lstms for target-dependent sentiment classification.
\newblock \emph{arXiv preprint arXiv:1512.01100}, 2015.

\bibitem[Vaswani et~al.(2017)Vaswani, Shazeer, Parmar, Uszkoreit, Jones, Gomez, Kaiser, and Polosukhin]{vaswani2017attention}
Ashish Vaswani, Noam Shazeer, Niki Parmar, Jakob Uszkoreit, Llion Jones, Aidan~N Gomez, {\L}ukasz Kaiser, and Illia Polosukhin.
\newblock Attention is all you need.
\newblock In \emph{Advances in neural information processing systems}, pp.\  5998--6008, 2017.

\bibitem[Wan et~al.(2020)Wan, Yang, Du, Liu, Qi, and Pan]{wan2020target}
Hai Wan, Yufei Yang, Jianfeng Du, Yanan Liu, Kunxun Qi, and Jeff~Z Pan.
\newblock Target-aspect-sentiment joint detection for aspect-based sentiment analysis.
\newblock In \emph{Proceedings of the AAAI Conference on Artificial Intelligence}, volume~34, pp.\  9122--9129, 2020.

\bibitem[Wang et~al.(2016)Wang, Huang, Zhu, and Zhao]{wang2016attention}
Yequan Wang, Minlie Huang, Xiaoyan Zhu, and Li~Zhao.
\newblock Attention-based lstm for aspect-level sentiment classification.
\newblock In \emph{Proceedings of the 2016 conference on empirical methods in natural language processing}, pp.\  606--615, 2016.

\end{thebibliography}
\bibliographystyle{iclr2025_conference}

\end{document}